\documentclass[conference]{IEEEtran}
\IEEEoverridecommandlockouts    
\usepackage{cite}
\usepackage{algorithm}
\usepackage{algorithmic}
\usepackage[algo2e]{algorithm2e}
\usepackage{amsmath,amssymb,amsfonts}
\usepackage{algorithm}
\usepackage{graphicx}
\usepackage{textcomp}
\def\BibTeX{{\rm B\kern-.05em{\sc i\kern-.025em b}\kern-.08em
T\kern-.1667em\lower.7ex\hbox{E}\kern-.125emX}}

\usepackage{lineno}
\usepackage{lettrine}

\usepackage{url}
\usepackage{moreverb}

\usepackage{graphicx}
\usepackage{subcaption}
\usepackage{float}
\usepackage{amsmath}
\usepackage{array}
\usepackage{multirow}
\graphicspath{ {Figures/} }

\usepackage{scalerel}
\usepackage{tikz}
\usetikzlibrary{svg.path}

\definecolor{orcidlogocol}{HTML}{A6CE39}
\tikzset{
	orcidlogo/.pic={
			\fill[orcidlogocol] svg{M256,128c0,70.7-57.3,128-128,128C57.3,256,0,198.7,0,128C0,57.3,57.3,0,128,0C198.7,0,256,57.3,256,128z};
			\fill[white] svg{M86.3,186.2H70.9V79.1h15.4v48.4V186.2z}
			svg{M108.9,79.1h41.6c39.6,0,57,28.3,57,53.6c0,27.5-21.5,53.6-56.8,53.6h-41.8V79.1z M124.3,172.4h24.5c34.9,0,42.9-26.5,42.9-39.7c0-21.5-13.7-39.7-43.7-39.7h-23.7V172.4z}
			svg{M88.7,56.8c0,5.5-4.5,10.1-10.1,10.1c-5.6,0-10.1-4.6-10.1-10.1c0-5.6,4.5-10.1,10.1-10.1C84.2,46.7,88.7,51.3,88.7,56.8z};
		}
}

\newcommand\orcidicon[1]{\href{https://orcid.org/#1}{\mbox{\scalerel*{
				\begin{tikzpicture}[yscale=-1,transform shape]
					\pic{orcidlogo};
				\end{tikzpicture}
			}{|}}}}

\usepackage{lipsum} 
\let\labelindent\relax
\usepackage{enumitem} 
\usepackage{booktabs}   
\usepackage{multirow}   

\usepackage{hyperref} 
\usepackage[nameinlink,noabbrev]{cleveref} 

\begin{document}
\title{
	\LARGE \bf
	\textit{From Prompts to Pavement Through Time:} Temporal Grounding in Agentic Scene-to-Plan Reasoning
}
    
\author{Ahmed~Y.~Gado$^{1,2}$, Omar~Y.~Goba$^{1,2}$, Alaa Hassanein$^3$ \\ Catherine~M.~Elias$^{1,2\orcidicon{0000-0002-1444-9816}\,}$,~\IEEEmembership{Member,~IEEE} and Ahmed Hussein$^{4}$,~\IEEEmembership{Senior Member,~IEEE}%
	\thanks{*This work was not supported by any organization}%
	\thanks{$^{1}$Computer Science \& Engineering Department, German University in Cairo (GUC), Egypt {\tt\small omaryasserassem1@gmail.com, ahmed.yahia.gado@gmail.com, catherine.elias@ieee.org} }%
	\thanks{$^{2}$C-DRiVeS Lab: Cognitive Driving Research in Vehicular Systems, Cairo, Egypt {\tt\small cdrives.researchlab@gmail.com}}%
	\thanks{$^3$ M.Eng. Robotics Candidate at Deggendorf Institute of Technology, Germany {\tt\small alaahg81@gmail.com}}%
	\thanks{$^{4}$IAV GmbH, Berlin, Germany {\tt\small ahmed.hussein@ieee.org}}%
}

\markboth{Journal of \LaTeX\ Class Files,~Vol.~14, No.~8, August~2015}%
{author1 \MakeLowercase{\textit{et al.}}:title here}
%



\maketitle

\begin{abstract}
	Recent attempts to support high-level scene interpretation and planning in Autonomous Vehicles (AVs) using ensembles of Large Language Models (LLMs) and Large Multimodal Models (LMMs) continue to treat time as a secondary property. This lack of temporal grounding leads to inconsistencies in reasoning about continuous actions, undermining both safety and interpretability. This work explores whether temporal conditioning within inter-agent communication can preserve or enhance coherence without introducing degradation in semantic or logical consistency. To investigate this, we introduce three planner architectures with progressively increasing temporal integration and evaluate them on curated subsets of the BDD-X dataset using semantic, syntactic, and logical metrics. Results show that while temporal conditioning reshapes reasoning style, it yields no statistically significant improvements in standard NLP-based correctness metrics. However, qualitative analysis reveals predictive hazard reasoning, stable corrective behavior, and strategic divergence in the Sentinel. These findings clarify the limits of prompt-based temporal grounding and establish the first empirical benchmark for temporal scene-to-plan reasoning.
\end{abstract}


%

\section{Introduction and State of Art}\label{sec1}

\IEEEPARstart{A}{utonomous} Vehicles (AVs) represent a transformative force in modern transportation, with the industry progressing through levels of automation defined by SAE International \cite{sae2021taxonomy}. The leap from Level 4 (L4) to Level 5 (L5) autonomy remains a formidable challenge, as it demands flawless operation in all possible conditions. Achieving this level of robustness hinges on sophisticated behavior planning, where the vehicle must generate a sequence of high-level actions based on its environment, goals, and, crucially, its understanding of events over time. A failure to account for this temporal dimension can lead to critical errors in tracking, prediction, and planning.

To address these complex reasoning tasks, the field has increasingly turned to Large Language and Multimodal Models (LLMs and LMMs), which are transformer-based architectures trained on vast datasets \cite{vaswani2017attention, dosovitskiy2020image}. These models have shown promise in various domains, including autonomous driving, where they can enhance perception, decision-making, and control capabilities\cite{huang2024robotron, tanahashi2023evaluation, cai2024driving}. Their powerful generalization capabilities can be further refined using techniques like In-Context Learning (ICL), which adapts model behavior through carefully crafted prompts and few-shot examples without requiring extensive retraining\cite{brown2020language, kojima2022large}. This adaptability has enabled the development of advanced agentic frameworks, or Multi-Agent Systems (MAS), where a swarm of specialized LMM agents can collaborate with each other and external tools to decompose and solve complex problems\cite{wang2024describe, xu2024magic}.

Following the promising results of MAS powered by a swarm of LMMs, the AV and robotics scientific community have explored extensively its application in their respective fields \cite{goba2026prompts, yu2025agentic, nisa2025agentic, nazar2025situational}. An overarching theme in the latest literature points to two primary axes of integration: perception and control. However, the temporal dimension, mentioned previously, remains absent from these recent advancements.

This absence constitutes a critical problem. Current agentic planners can respond to instantaneous stimuli but lack the ability to maintain causal continuity across consecutive actions. As a result, decisions that appear coherent locally may become inconsistent over time, undermining both safety and long-horizon interpretability.

While prior research has enhanced the perceptual and control capacities of agentic systems, the reasoning layer; where temporal alignment should emerge, remains underdeveloped. No current architecture explores the effects of temporal grounding as a first-class property in multi-agent planning.

This work explores whether temporal conditioning within inter-agent communication can preserve or enhance coherence without introducing degradation in semantic or logical consistency. In particular, we examine whether defining planner–executor interactions as temporally conditioned message exchanges allows a swarm of LMM agents to retain contextual awareness of prior states without explicit memory mechanisms.

\begin{figure*}[t]
	\centering
	\begin{subfigure}{0.32\textwidth}
		\includegraphics[width=\linewidth]{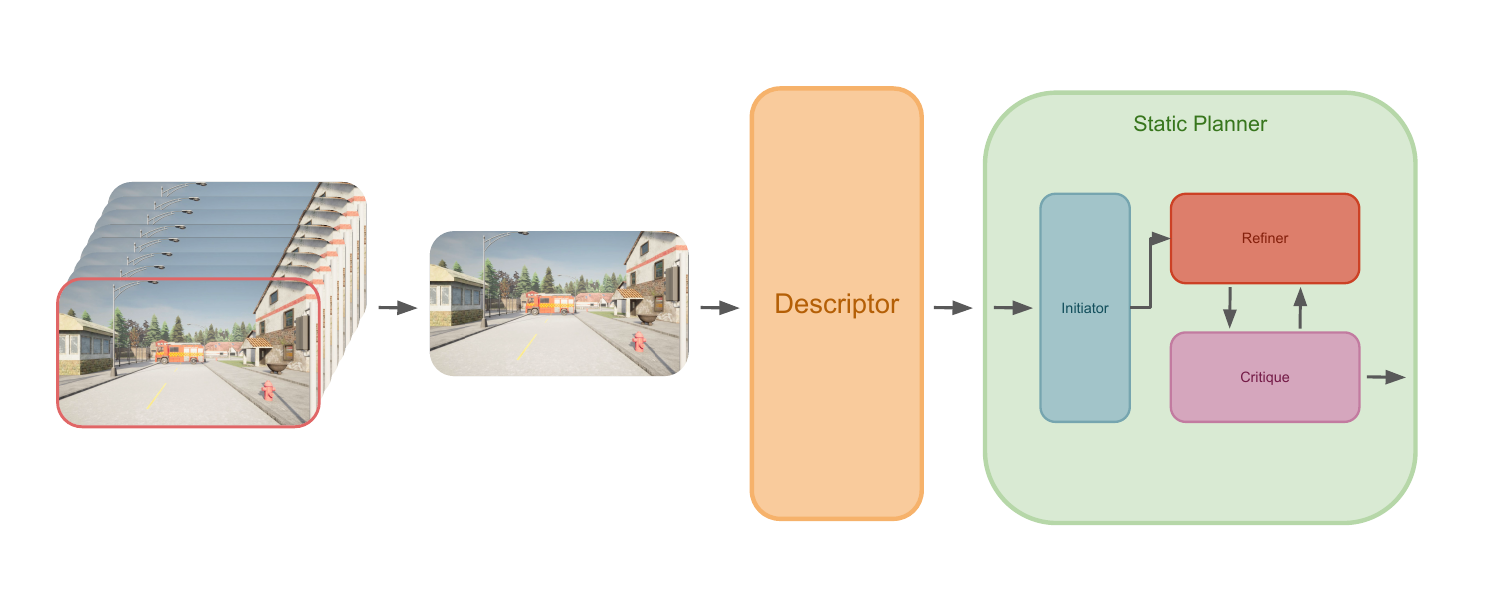}
		\caption{Static Planner}
		\label{fig:static}
	\end{subfigure}\hfill
	\begin{subfigure}{0.32\textwidth}
		\includegraphics[width=\linewidth]{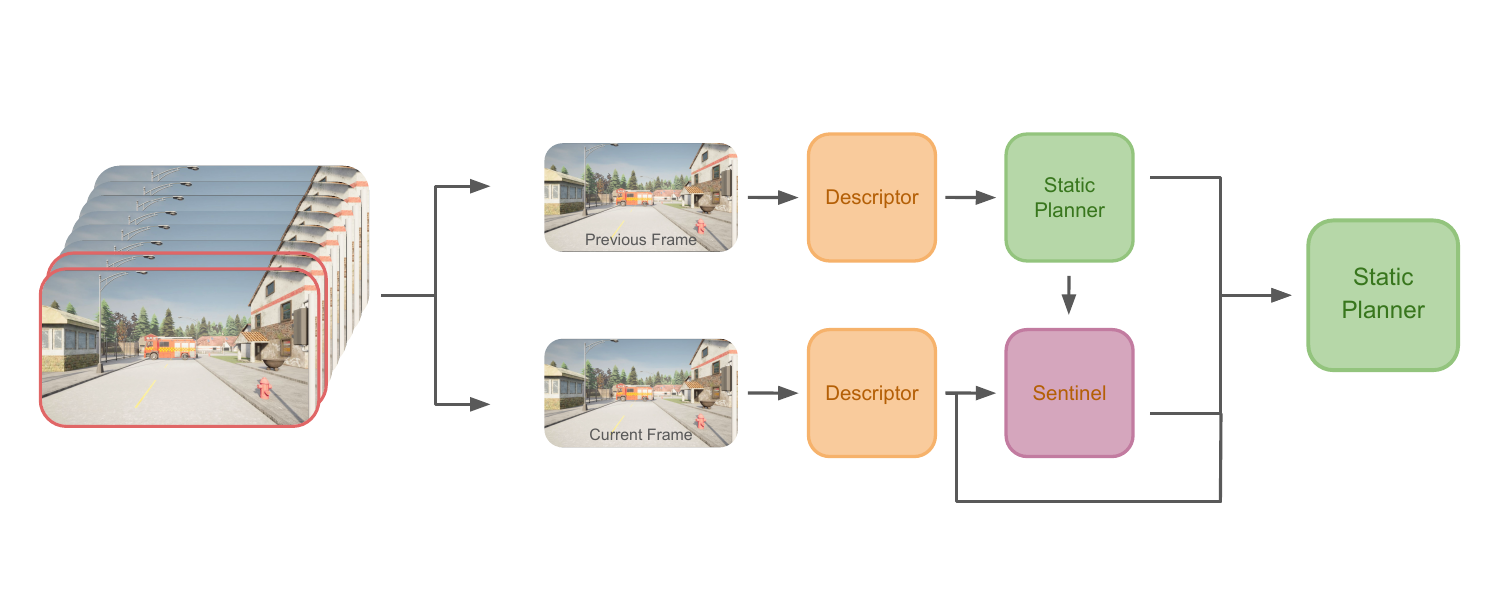}
		\caption{Sentinel Planner}
		\label{fig:sentinel}
	\end{subfigure}\hfill
	\begin{subfigure}{0.32\textwidth}
		\includegraphics[width=\linewidth]{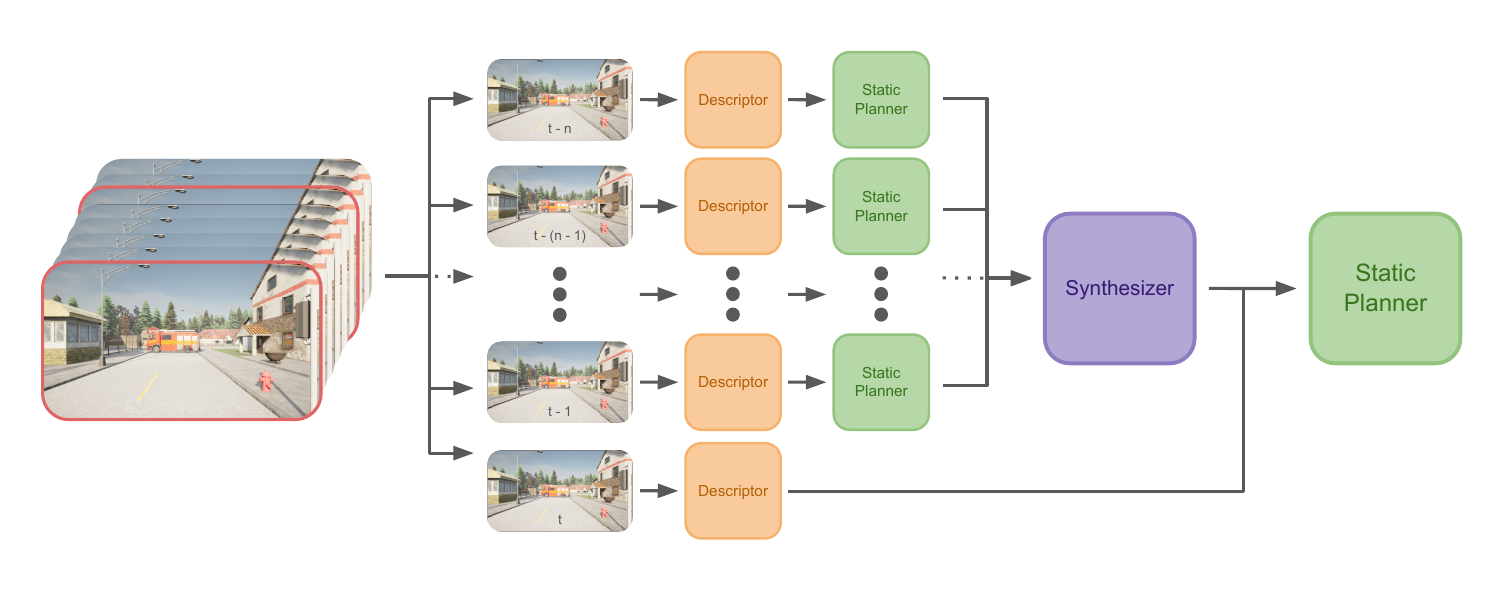}
		\caption{Synthesizer Planner}
		\label{fig:synthesizer}
	\end{subfigure}

	\caption{Overall system architectures of the three planners with increasing temporal grounding.}
	\label{fig:architectures}
\end{figure*}
To examine this hypothesis, the contributions of this work are organized along three complementary dimensions:

\begin{enumerate}
	\item A temporal extension to the \textit{Prompt-to-Pavement} \cite{goba2026prompts} formulation that introduces multi-frame contextual reasoning, enabling planners to maintain temporal coherence across sequential decisions.

	\item A novel evaluation metric for logical continuity between successive plans, capturing temporal–causal coherence beyond surface-level semantic similarity.

	\item A comparative study of static, one-step temporal, and multi-frame temporal architectures for LMM-driven behavior planning, establishing a reproducible benchmark for temporal grounding in AVs.
\end{enumerate}

\section{Methodology}\label{sec3}

Section \ref{sec1} showed that current MAS have trouble with
understanding time in changing environments like AVs. To fix this break in
understanding, we introduce a series of planners that grow from seeing just one
moment to thinking in terms of a story, with each planner building on the one
before it and adding more layers of temporal understanding. Figure
\ref{fig:architectures} gives a high level summary of the three planner systems
introduced in this section: (a) the Static Planner works only with the current
observation, (b) the Sentinel Planner adds short term time alignment, and (c)
the Synthesizer Planner includes long term memory of the situation
\vspace{-5pt}
\subsection{The Agentic Backbone}\label{sec3a}
At the center of our method is an agent-based backbone that was built based on
the "From Prompts to Pavement" framework \cite{goba2025prompts2pavement}. The system
is made of three main parts: \textbf{The Descriptor}, \textbf{The
	Planner}, and \textbf{The Generator}. Each part, which is based on an LMM,
is specialized for its role, and they work together to let the vehicle
understand, plan, and move through its environment in new situations.

\subsubsection{The Descriptor Agent}\label{sec:descriptor}
This agent acts as the first step in the process. It takes an image $i$ and creates a detailed textual description of the scene, structured as an observation $o = \langle \kappa, e, d \rangle$, where $\kappa$ is a criticality key, $e$ is an issue explanation, and $d$ is the scene description. To improve its spatial reasoning, we use a Chain-of-Symbols (CoS) approach \cite{chain-of-symbols}. This agent is unchanged from the original framework \cite{goba2025prompts2pavement}.

\subsubsection{The Planner Agent}\label{sec:planner}
This agent takes the observation $o$ from the Descriptor and creates a high-level plan $p = [g_1, g_2, \ldots, g_n]$. This plan is an ordered sequence of goals ($g_k \xrightarrow{order} g_{k+1}$) designed to resolve the identified issue, meaning the actions must be done in a specific sequence to get the wanted result.

\subsubsection{The Generator Agent}\label{sec:generator}
The final Generator agent, which created executable behavior trees, was removed from this study to focus purely on the planning stage.
These three parts make up the backbone of our agent system. The
next sections will explain the improvements made only to the planner to add abilities for thinking about time.
\vspace{-5pt}
\subsection{The Static Planner}\label{sec3b}
The Static Planner serves as our non-temporal baseline, operating only on the current observation. Its design is inspired by the actor-critic model and our prior work \cite{goba2025prompts2pavement, konda1999actor}, using a multi-agent refinement process shown in Figure~\ref{fig:architectures}a.

\subsubsection{The Initiator Agent}\label{sec:initiator}
This agent creates an initial, high-level plan $p_0$ based only on the current observation $o$. This initial plan is created greedily and requires further refinement.

\subsubsection{The Critic Agent}\label{sec:critic}
The Critic evaluates the plan $p_i$ for mistakes, producing an output $c = \langle \texttt{HALT}, f \rangle$. If the plan is acceptable, the $\texttt{HALT}$ signal is set to 1; otherwise, the string $f$ contains feedback for improvement.

\subsubsection{The Refiner Agent}\label{sec:refiner}
If refinement is needed, the Refiner takes the current plan $p_i$ and the critic's feedback $c$ to produce an improved plan, $p_{i+1}$. This loop continues until the plan is accepted or a maximum number of steps is reached. This iterative process ensures that any performance gains in other planners can be credited to their use of time-based information, not just the refinement process itself.

\vspace{-5pt}
\subsection{The Sentinel Planner}\label{sec3c}
To introduce the idea of time in a step-by-step way, we first create the
Sentinel Planner, shown in Figure~\ref{fig:architectures}b, which extends
the Static baseline by adding a short-term memory system that holds
information from the most recent past moment. This addition, we hypothesize,
will allow the planner to maintain continuity across back-to-back observations,
addressing the time discontinuity problem identified earlier.

The \textbf{Sentinel Planner} keeps the same three-agent structure as the
Static Planner: the Initiator, the Critic, and the Refiner. However, there is a
new agent, named \textbf{the Sentinel}. This agent is responsible for
analyzing the match between the current description being made by the
Descriptor and the previous plan that was generated. The Sentinel's job is to make sure
that the new plan corrects any differences from the previous one by using
the current observation as context. Formally, the Sentinel is defined as:
\[
\textbf{Sentinel}\left(
	o^{(t)},
	p^{(t-1)},
	\texttt{FPS},
	\Delta t
	\right) = s \in \mathcal{S}
	~s.t.~
\]
\[
	o^{(t)} \in \mathcal{O},~
	p^{(t-1)} \in \mathcal{G}^+,~
	\texttt{FPS} \in \mathbb{Z}^+,~
	\Delta t \in \mathbb{R}^+
\]
\[
	s \equiv \left\langle
	\texttt{ALIGNMENT} \in \{0,~1\},
	j \in \Sigma^+
	\right\rangle
\]

Where $o^{(t)}$ is the current observation at time $t$, and
$p^{(t-1)}$ is the plan made in the previous time step. In addition
to these inputs, the Sentinel receives the frame rate (\texttt{FPS}) and the
time passed between steps ($\Delta t$), which together give more
time-based context. Using this information, the Sentinel judges whether the
current state of the world is still consistent with what the agent intended before. It
outputs an alignment signal and a justification string that
explains the reason for this judgment. When the \texttt{ALIGNMENT} signal
is 1, the current observation is considered consistent with the previous plan;
otherwise, the justification string points out the source of the mismatch and
signals that a corrective improvement is needed.

In short, the Sentinel keeps the agent on track at each time step by checking
if the current world state still matches what the agent previously
wanted. However, this system only provides local alignment. In sequential
planning, small differences that do not break the immediate goal can
still change future possibilities. This means that small errors can build
up over time. So, short-term checking is a step in the right
direction but is not enough for keeping the overall task goal. To do that,
the planner must include a formal way of understanding time over the long term.

\vspace{-5pt}
\subsection{The Synthesizer Planner}\label{sec3d}
To fully solve the time-understanding challenge, the introduction of
\textbf{the Synthesizer Planner}, shown in Figure~\ref{fig:architectures}c,
is needed. This planner builds on the Sentinel Planner by adding a
long-term memory system that holds information across many past
time steps. This, in theory, allows the planner to maintain a consistent story
over time, making sure that each plan matches not only the immediate past but
also the larger series of events.

The Synthesizer Planner keeps the same agent structure as the Static
planner, with the addition of a new agent called \textbf{the Synthesizer}. This agent is
responsible for creating a complete understanding of the vehicle's
history over time. To get this understanding, the Synthesizer builds a
story over time by combining a series of both past observations and plans. Formally,
\[
	\textbf{Synthesizer}\left(
	o^{(t-k:t-1)},
	p^{(t-k:t-1)},
	o^{(t)}
	\right) = n \in \Sigma^+
	~s.t.~
\]
\[
	o^{(t-k:t-1)} \in \mathcal{O}^+,~
	p^{(t-k:t-1)} \in \mathcal{G}^{+},~
	o^{(t)} \in \mathcal{O}
\]
Where $o^{(t-k:t-1)}$ is the series of past observations from
time step $t-k$ up to the previous time step $t-1$, and $p^{(t-k:t-1)}$
is the matching series of plans made during that same time.
These paired series capture how the environment has changed and
how the agent has previously reacted to those changes. The Synthesizer also
takes the current observation $o^{(t)}$ as input, connecting the reconstructed
story over time to the present situation. By combining both historical
observations and past decision records, the Synthesizer builds a
\textit{narrative representation}, shown as $n$, which is a structured
text story that holds the cause-and-effect and time-based relationships
seen over time. This story serves as a high-level summary of the
agent's past experience, encoding patterns of continuity, difference, and
intent. The resulting representation is then given to the planner’s three
component agents, allowing them to think not only about the immediate scene
but also about how it has developed. In doing so, the Synthesizer grounds the
planner’s thinking in a time-based context, allowing for more consistent and
forward-looking decision-making that reflects both the vehicle’s past behavior and
its expected path.

Having set up the three planners, the following section explains the empirical
basis for their evaluation: the datasets, sampling methods, and measurements
that make temporal consistency observable and measurable.

\section{Dataset and Evaluation Setup}\label{sec4}

As mentioned before, the evaluation process was designed to capture the
benefits of adding time-based understanding in steps within planning tasks.
For this reason, the performance of our three proposed planners is checked across
four key areas: Syntactic Correctness, Semantic Alignment, Logical
Coherence, and Generation Efficiency. To properly evaluate these
areas, the chosen dataset must contain time-based changes that allow a
full examination of each planner’s reasoning abilities.

The dataset used for evaluation is the \textit{Berkeley DeepDrive} (BDD-X)
dataset \cite{kim2018textual}. It comprises 16,394 driving videos, each paired
with human-written labels detailing driver actions and their rationale.
BDD-X is well-suited for this study due to its rich time-based information,
linking video frames with textual explanations of driver decisions, enabling
accurate measurement of a planner’s ability to create temporally consistent plans.

To find the hidden time-based continuity in this otherwise broken-up collection of data,
we use two methods that work together: (1) a targeted subset creation step,
and (2) a random sampling strategy. Together, these methods ensure both
control over the meaning and variety in the timing during the evaluation process.
\vspace{-5pt}
\subsection{Targeted Subset Creation}\label{sec4.1}
To make the evaluation results easier to understand, a collection of
twelve targeted subsets was manually created from the original BDD-X dataset,
with each one having a different size based on how many relevant samples there were. The
process of grouping the data was done with help from an LLM, which was
guided to group action labels into categories that made sense together. Each
subset was designed around a specific driving theme or situation, such
as cycling, pedestrian crossings, road blocks, or traffic signals. The
resulting subsets together cover all videos in the dataset, making sure that
each video is included in at least one subset. Each created subset thus
forms a controlled setting where thinking about time can be evaluated under
different specific conditions.
\vspace{-5pt}
\subsection{Stochastic Sampling Method}\label{sec4.2}
Beyond the manually created subsets, a random sampling method was used
to show variety in timing that is not captured by the controlled meaning-based conditions.
In real driving, events that trigger a response appear without warning, requiring the MAS
to respond by creating a plan within a very short time context. To copy
this characteristic, frames were randomly picked from each video in the dataset,
with each frame acting as an independent time-based signal that starts the plan
creation process. Through this method, the static BDD-X videos are re-read
as a series of separate inputs, with each one representing a moment of
decision-making rather than a continuous path. The resulting setup
creates a dynamic testing environment, allowing planners to be checked under
scenarios that are broken up in time but still connected by cause and effect.
\vspace{-5pt}
\subsection{Evaluation Metrics}\label{sec4.3}
To measure the performance of each planner, standard evaluation metrics were
employed across four key areas. Syntactic Correctness was assessed using METEOR
\cite{banerjee2005meteor} and ROUGE \cite{lin2004rouge} for word-based
similarity. Semantic Alignment was evaluated with BERTScore
\cite{zhang2019bertscore}, which calculates contextual similarity via BERT
embeddings. Logical Coherence was determined using a Natural Language Inference
(NLI) technique with \textit{bart-large-mnli} \cite{lewis2019bart}, focusing on
entailment probability. Generation Efficiency was measured by total generation
time and token cost. These metrics collectively provide a comprehensive
assessment of the planners' abilities in time-based planning tasks.
\vspace{-5pt}
\subsection{Statistical Testing}\label{sec4.4}

The reason for using statistical significance testing here is to ensure that
the observed performance differences between planners are not because of
random variation. To do that, the following hypotheses were created:

\begin{enumerate}[label=\textbf{H\arabic*}, ref=H\arabic*]
	\item \label{hyp:H1} The Static Planner performs worse than the Sentinel Planner on the semantic alignment and logical coherence metrics.
	\item \label{hyp:H2} The Static Planner performs worse than the Synthesizer Planner on the same metrics.
	\item \label{hyp:H3} The Sentinel Planner performs worse than the Synthesizer Planner on the same metrics.
\end{enumerate}

Together, these hypotheses define a direct performance ordering across the
three planners on the semantic and logical areas. Both syntactic correctness
and generation efficiency were left out, as they do not reflect the level of
time-based understanding in planning behavior.

Before doing the statistical test, the raw measurement outputs were changed
using a heavy-sided log transform. This change was done because of the severe
score compression seen in semantic similarity measurements, where values group
together near the upper limit. The heavy-sided version of the log transform
increases sensitivity in the high-scoring region by making small differences
larger that would otherwise remain statistically impossible to tell apart. The
transformed values were used for all statistical calculations, while the raw
scores were kept for reporting.
\[
	t(s)=
	\begin{cases}
		-\log(1-s),        & \epsilon \le s \le 1-\epsilon \\
		-\log(\epsilon),   & s > 1-\epsilon                \\
		-\log(1-\epsilon), & s < \epsilon
	\end{cases}
\]

The statistical test chosen was the permutation test. It is a non-parametric
method that does not assume any specific underlying statistical distribution. This
choice is very important since transformer-based models produce complex, context-
based distributions that do not follow standard Gaussian assumptions. The permutation
test works on the paired mean differences between planner versions
($\Delta$-means). With three planners, six pairwise comparisons were checked.
For each metric, the $\Delta$-means were calculated independently, and the signs
of these differences were randomly flipped for each permutation, repeated
$10^5$ times to build the null distribution. The resulting $p$-value
matches the proportion of permuted $\Delta$-means with sizes
greater than or equal to the observed one.

To add to the $p$-value, the effect size was estimated using Cohen’s $d$,
and a 95\% confidence interval was calculated through bias-corrected accelerated
(BCa) bootstrapping. The bootstrap sample size was set to $10^4$ iterations.
Furthermore, to account for multiple comparisons across all planner pairings
and metrics, a Benjamini–Hochberg correction was applied to control the false
discovery rate. The combination of the heavy-sided transform, permutation-based
testing, bootstrapped intervals, and adjusted $p$-values ensures that the
reported differences are both statistically trustworthy and practically
easy to understand.

\section{Results and Discussion}\label{sec5}

The analysis opens by restating the core hypothesis: \textit{that temporal conditioning within inter-agent communication can sustain or enhance coherence without degrading semantic or logical consistency.} The performance of three planners, \textbf{Static}, \textbf{Sentinel}, and \textbf{Synthesizer}, is compared across five semantic groups from the BDD-X dataset, which in total is approximately 1300 videos. Each planner was run using 4 frames per video. These frames were randomly sampled (as explained in Section \ref{sec4}), but they were limited to the first 2 seconds of the video. This restriction was put in place to make sure the model could not rely on foresight by seeing too much of the future. Two groups, \texttt{weather\_condition} and \texttt{pedestrian\_crossing}, create controlled sets for direct comparison across models, while three others, \texttt{lane\_merge\_or\_turn}, \texttt{oncoming\_traffic}, and \texttt{pedestrian\_crossing}, are used to study the Sentinel’s thinking behavior by itself.

The evaluation covers semantic alignment (BERTScore), logical consistency (NLI), word-based similarity (ROUGE, METEOR), and computational efficiency (generation time and token cost). Together, these measurements are used to capture the coherence, accuracy, and efficiency when using time-based guidance.

The next subsections report the measured results, showing how the depth of time-based information influences coherence, efficiency, and language cost.
\vspace{-5pt}

\subsection{Quantitative Evaluation}
The quantitative results, which are summarized in Tables \ref{tab:comprehensive_results} and \ref{tab:subsets_comprehensive_results}, show that while small changes exist across word-based, semantic, and logical measurements, no planner shows a consistent or clear advantage. A clear difference only appears for computational efficiency, where the \textbf{Sentinel} proves to be the most token-efficient time-aware version. However, these small differences are not statistically important. As shown in Figure~\ref{fig:statistical_analysis}, p-values remained well above the 0.05 threshold, and Cohen's d effect sizes were consistently weak ($|d| < 0.2$). This confirms that, based on the numbers, the null hypothesis is correct. These quantitative trends show the efficiency and coherence profiles of each planner, but they only tell part of the story.
\vspace{-10pt}
\begin{table}[H]
	\centering
	\caption{Comprehensive Performance Metrics. Mean$\pm$std. Best values in bold: Across Planner Architectures}
	\label{tab:comprehensive_results}
		\scriptsize
		\setlength{\tabcolsep}{2pt}
		\begin{tabular}{|p{1.55cm}|p{1.55cm}|p{1.55cm}|p{1.55cm}|p{1.55cm}|}
			\hline
			\textbf{Metric}     & \textbf{Category} & \textbf{Sentinel}        & \textbf{Synthesizer}     & \textbf{Vanilla}           \\ \hline
			\multicolumn{5}{|c|}{\textit{Semantic \& Lexical Coherence}} \\ \hline
			ROUGE-L & Justifications& 0.081$\pm$0.051          & \textbf{0.082$\pm$0.047} & 0.079$\pm$0.050  \\
			  & Plans & 0.058$\pm$0.060 & 0.061$\pm$0.052 & \textbf{0.062$\pm$0.053}  \\ \hline
			METEOR & Justifications & \textbf{0.120$\pm$0.052} & 0.117$\pm$0.049 & 0.118$\pm$0.052 \\
			& Plans & 0.066$\pm$0.073 & \textbf{0.067$\pm$0.063} & 0.062$\pm$0.056 \\ \hline
			BERTScore (F1) & Justifications & \textbf{0.855$\pm$0.010} & 0.853$\pm$0.011 & \textbf{0.855$\pm$0.010}   \\
			& Plans & 0.839$\pm$0.013 & \textbf{0.840$\pm$0.014} & 0.835$\pm$0.071 \\ \hline
			\multicolumn{5}{|c|}{\textit{Logical Consistency (NLI)}} \\ \hline
			Entailment & Justifications & 0.039$\pm$0.119 & \textbf{0.046$\pm$0.117} & 0.039$\pm$0.104 \\
			& Plans & 0.082$\pm$0.097 & \textbf{0.098$\pm$0.153} & 0.096$\pm$0.129 \\ \hline
			Neutral & Justifications & 0.859$\pm$0.279          & 0.861$\pm$0.268 & \textbf{0.869$\pm$0.262} \\
			& Plans & \textbf{0.904$\pm$0.118} & 0.832$\pm$0.233 & 0.849$\pm$0.204 \\ \hline
			Contradiction & Justifications & 0.102$\pm$0.266 & \textbf{0.093$\pm$0.254} & \textbf{0.093$\pm$0.251} \\
			& Plans & \textbf{0.014$\pm$0.055} & 0.070$\pm$0.198 & 0.055$\pm$0.176 \\ \hline
			\multicolumn{5}{|c|}{\textit{Computational Efficiency}} \\ \hline
			Generation Time (s) & -- & 33.396$\pm$44.635        & 40.720$\pm$51.110 & \textbf{20.960$\pm$35.381} \\ \hline
			Token Cost & -- & \textbf{58975$\pm$13115} & 64556$\pm$20488 & 61729$\pm$21630 \\ \hline
		\end{tabular}
\end{table}
\vspace{-25pt}
\begin{table}[H]
	\centering
	\caption{Comprehensive Performance Metrics. Mean$\pm$std. Best values in bold: Across Traffic Scenarios}
		\label{tab:subsets_comprehensive_results}
		\scriptsize
		\setlength{\tabcolsep}{2pt}
		\begin{tabular}{|p{1.55cm}|p{1.55cm}|p{1.55cm}|p{1.55cm}|p{1.55cm}|}
			\hline
			\textbf{Metric}     & \textbf{Category} & \textbf{Pedestrian}      & \textbf{Oncoming}         & \textbf{Merge/Turn} \\ \hline
			\multicolumn{5}{|c|}{\textit{Semantic \& Lexical Coherence}}                                                         \\ \hline
			ROUGE-L             & Justifications    & \textbf{0.100$\pm$0.049} & 0.058$\pm$0.049           & 0.058$\pm$0.046     \\
			                    & Plans             & 0.094$\pm$0.062          & \textbf{0.098$\pm$0.050}  & 0.083$\pm$0.053     \\ \hline
			METEOR              & Justifications    & \textbf{0.139$\pm$0.071} & 0.095$\pm$0.047           & 0.102$\pm$0.050     \\
			                    & Plans             & \textbf{0.107$\pm$0.077} & 0.102$\pm$0.061           & 0.093$\pm$0.063     \\ \hline
			BERTScore (F1)      & Justifications    & \textbf{0.855$\pm$0.011} & 0.848$\pm$0.008           & 0.844$\pm$0.009     \\
			                    & Plans             & \textbf{0.847$\pm$0.030} & 0.846$\pm$0.014           & 0.843$\pm$0.035     \\ \hline
			\multicolumn{5}{|c|}{\textit{Logical Consistency (NLI)}}                                                             \\ \hline
			Entailment          & Justifications    & \textbf{0.033$\pm$0.085} & 0.032$\pm$0.053           & 0.034$\pm$0.083     \\
			                    & Plans             & \textbf{0.109$\pm$0.198} & 0.090$\pm$0.137           & 0.089$\pm$0.171     \\ \hline
			Neutral             & Justifications    & \textbf{0.854$\pm$0.274} & 0.823$\pm$0.295           & 0.809$\pm$0.315     \\
			                    & Plans             & \textbf{0.765$\pm$0.317} & 0.754$\pm$0.309           & 0.713$\pm$0.353     \\ \hline
			Contradiction       & Justifications    & \textbf{0.113$\pm$0.271} & 0.145$\pm$0.302           & 0.157$\pm$0.317     \\
			                    & Plans             & \textbf{0.126$\pm$0.289} & 0.155$\pm$0.315           & 0.198$\pm$0.351     \\ \hline
			\multicolumn{5}{|c|}{\textit{Computational Efficiency}}                                                              \\ \hline
			Generation Time (s) & --                & \textbf{76.84$\pm$24.10} & 98.27$\pm$70.27           & 101.32 $\pm$90.31    \\ \hline
			Token Cost          & --                & 238772 $\pm$31410         & \textbf{233264$\pm$20477} & 237013$\pm$29952    \\ \hline
		\end{tabular}
\end{table}
\vspace{-20pt}
\subsection{Exploratory Profile of the Sentinel Planner}
The \textbf{Sentinel} was chosen for more analysis because of its better token efficiency, which marks it as the most scalable time-aware version. As shown in Figure \ref{fig:radar} and Table \ref{tab:subsets_comprehensive_results}, the Sentinel maintains a stable and balanced performance across different traffic situations. Crucially, it does this with coherence that is close to the baseline, showing only small differences ($\delta \approx 0.01-0.02$ in BERTScore) compared to the static planner. This supports the claim that it is not worse, where meaning over time is preserved within the normal range of error while also reducing the amount of computation needed.
\vspace{-10pt}
\subsection{Qualitative and Analytical Reflection}
\begin{figure}[H]
	\centering
	\includegraphics[width=0.8\linewidth]{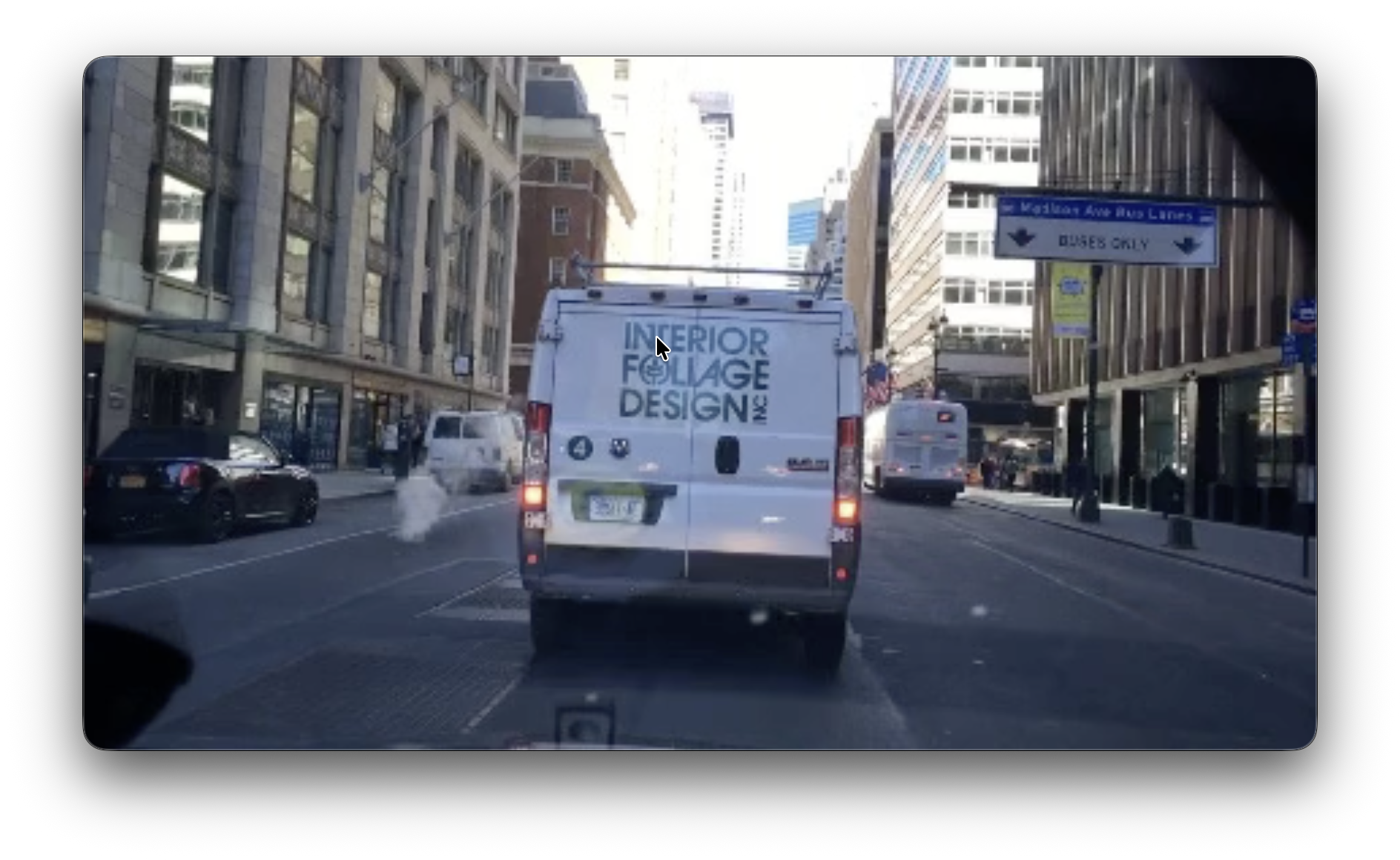}
	\caption{An example of the model outsmarts the ground truth—spotting a pedestrian and stopping early while the annotation fixates on a bus. Next frame? The “truth” catches up, proving the model was right all along.}
	\label{fig:qualitative_example}
    \begin{tabular}{p{0.25\linewidth} p{0.65\linewidth}}
		\toprule
		\textbf{Ground Truth} & \textbf{Model Prediction} \\
		\midrule
		\textit{Justification:} since there no way to move faster ahead. & \textit{Justification:} This plan effectively addresses the need to maintain legal compliance in lane usage while ensuring readiness for dynamic traffic conditions, prioritizing safety and awareness. \\
		\addlinespace
		\textit{Plan:} The car moves forward slowly                      & \textit{Plan:} Maintain steady speed while monitoring the lane ahead for constraints. Be prepared to execute a lane change if necessary when the bus maneuvers.                                         \\
		\bottomrule
	\end{tabular}
\end{figure}

Beyond the number-based measurements, a quality-based analysis reveals a more complex story. The model's differences from the ground truth are not usually errors, but are evidence of advanced thinking. We see a pattern of \textit{Predictive Hazard Identification}, where the model actively finds and reacts to critical safety events, such as a pedestrian entering a crosswalk, often before this is noted in the slow-to-update ground-truth annotation. This leads to \textit{Purposeful Strategic Divergence}, as illustrated in Figure \ref{fig:qualitative_example}, where the model creates a valid, and often safer, alternative plan that is different from the incomplete ground truth. Furthermore, its unique justification style—which is more descriptive and abstract than the dataset's direct cause-and-effect statements—is a good sign of generalization through in-context learning. This shows the model is re-thinking the scene rather than just copying the annotation style. Because of this, the combination of low word-based similarity (ROUGE, METEOR) with high meaning-based coherence (BERTScore) is not a conflict, but is a strong positive sign of successful generalization.
\vspace{-7pt}
\subsection{Interpreting the Null Hypothesis: The Limits of Prompting and Metrics}
The lack of statistically important improvement is a useful finding in itself, marking the boundary where simple prompt-based guidance ends. We believe this null result is due to two main factors. First, and most critically, is the built-in lack of detail in the ground-truth dataset. 
\begin{figure}[H]
	\centering
	\begin{subfigure}{0.8\linewidth}
		\includegraphics[width=\linewidth]{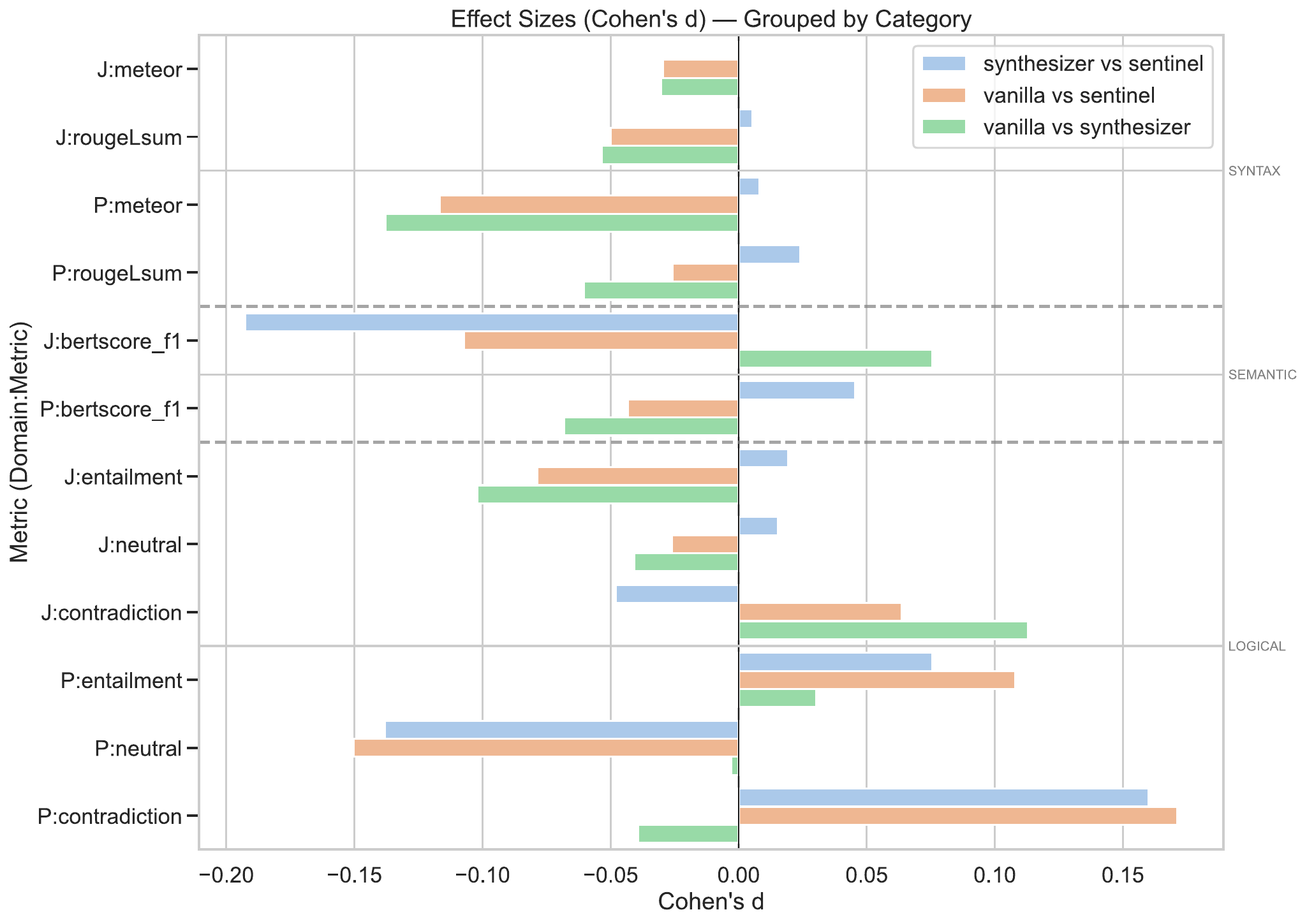}
		\caption{Cohen's d}
	\end{subfigure}
	\begin{subfigure}{0.8\linewidth}
		\includegraphics[width=\linewidth]{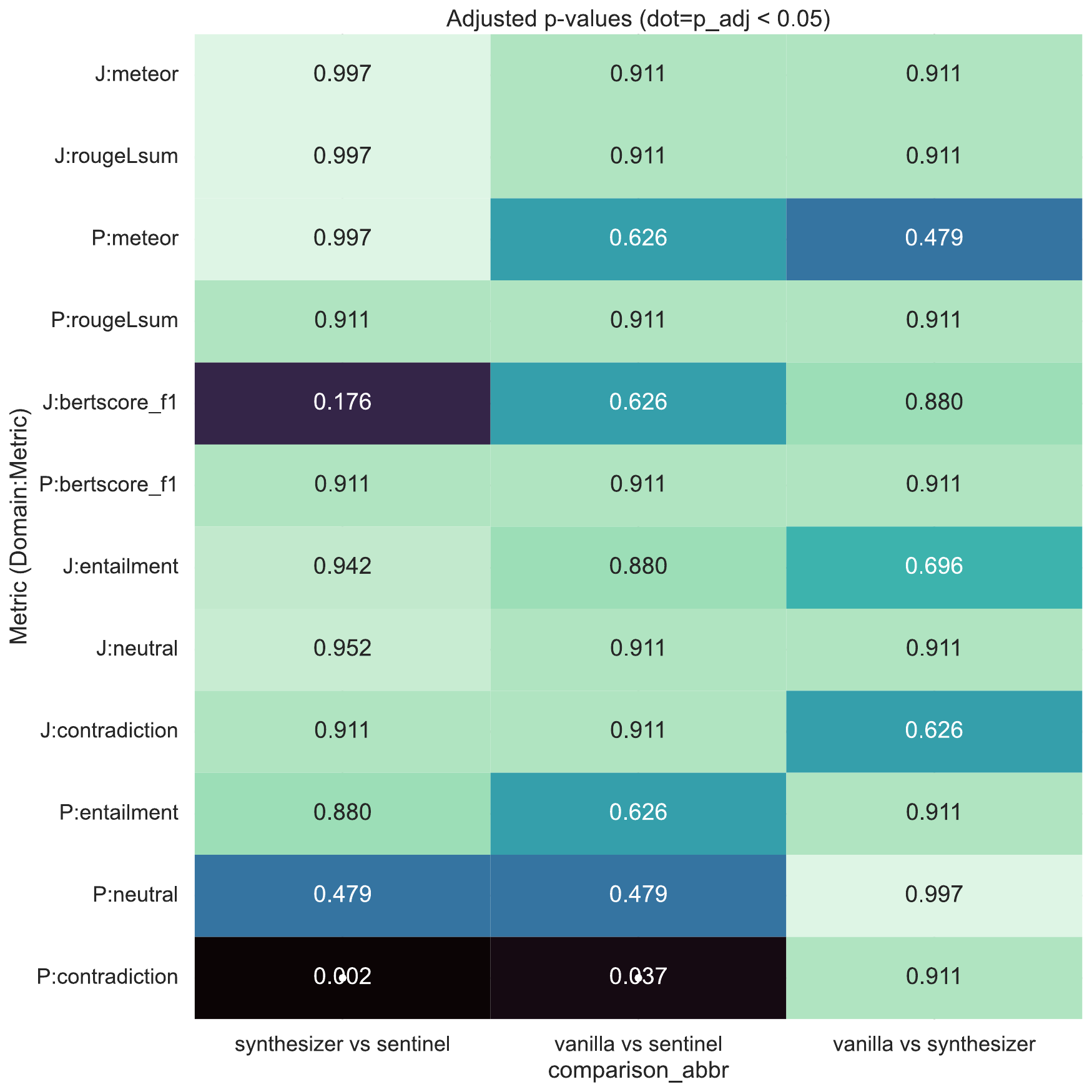}
		\caption{P-value heatmap}
	\end{subfigure}
    \end{figure}
    \begin{figure}[H]\ContinuedFloat
    \centering
    \begin{subfigure}{0.65\linewidth}
		\includegraphics[width=\linewidth]{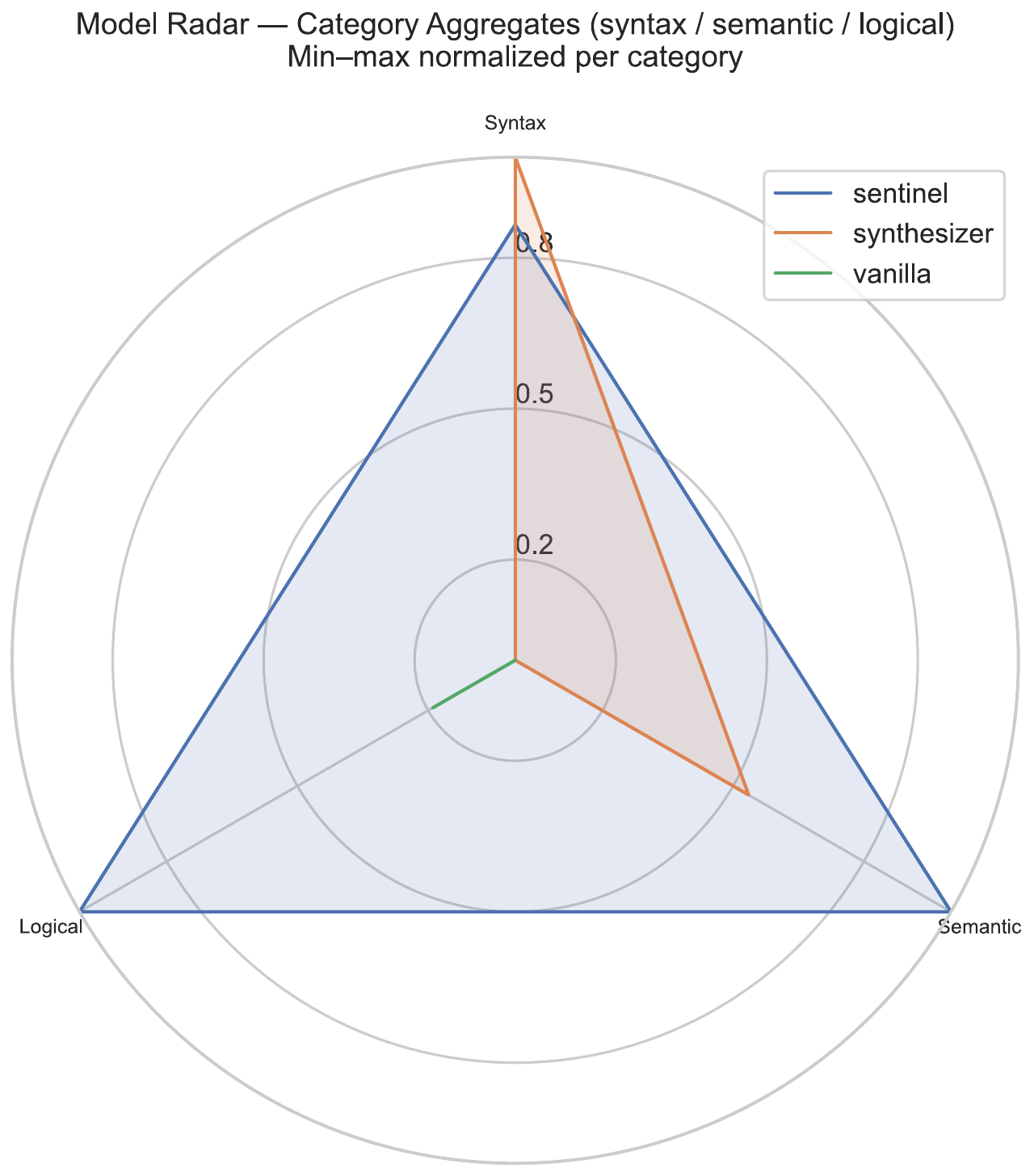}
		\caption{Radar plot}
        \label{fig:radar}
	\end{subfigure}
    \begin{subfigure}{0.65\linewidth}
		\includegraphics[width=\linewidth]{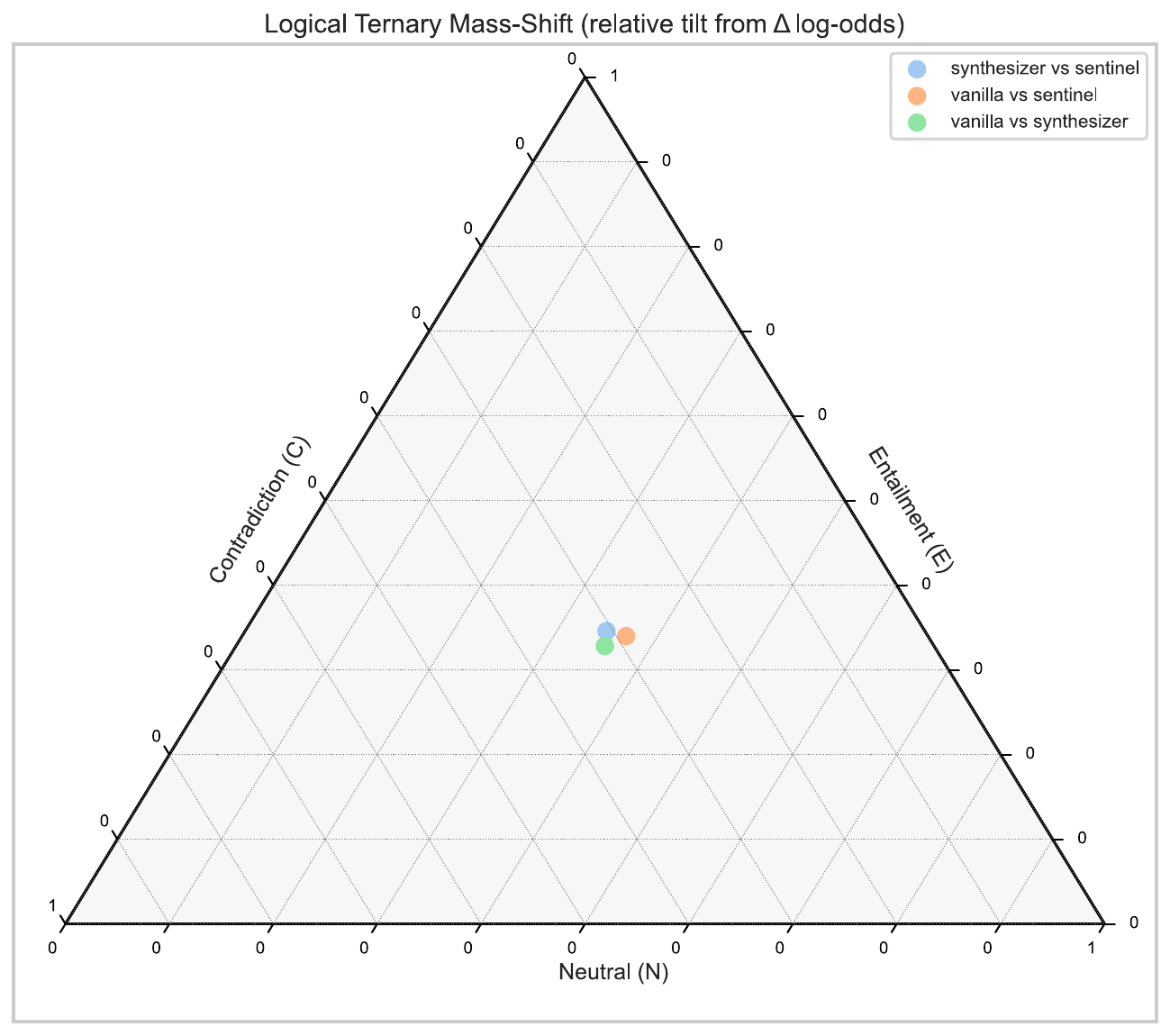}
		\caption{Ternary plot}
	\end{subfigure}
    \begin{subfigure}{0.8\linewidth}
		\includegraphics[width=\linewidth]{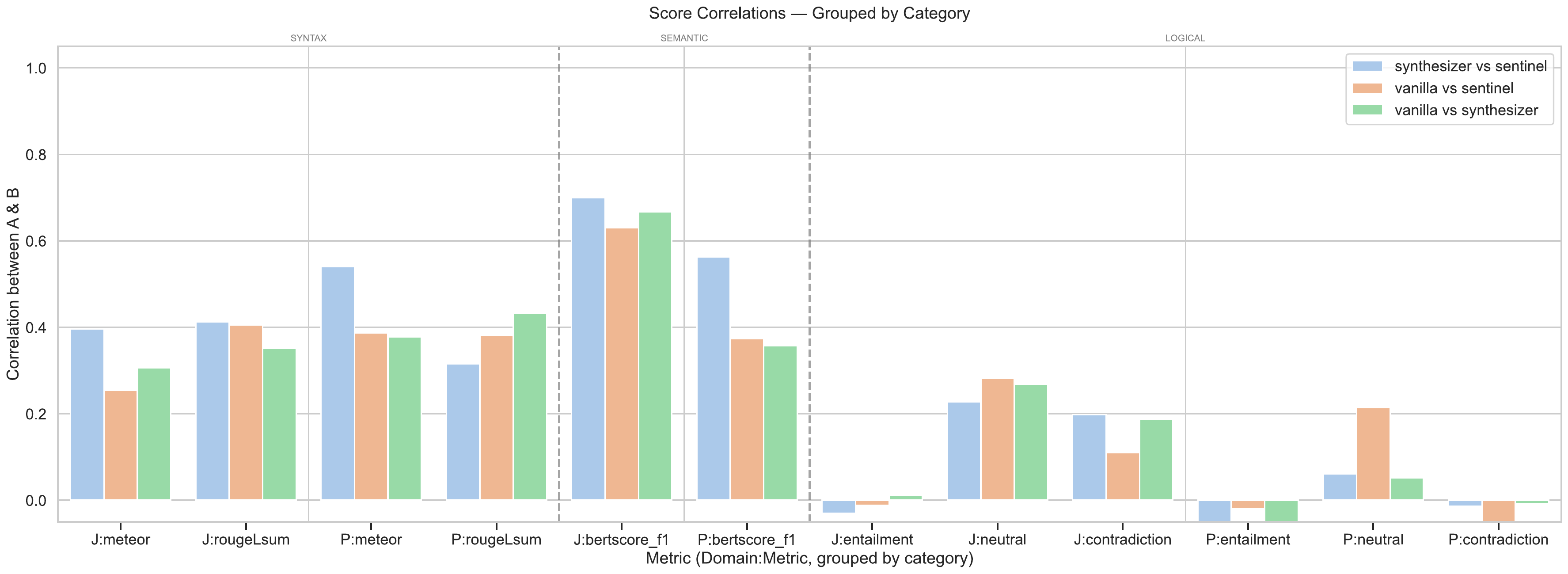}
		\caption{Correlations}
	\end{subfigure}
	\begin{subfigure}{0.8\linewidth}
		\includegraphics[width=\linewidth]{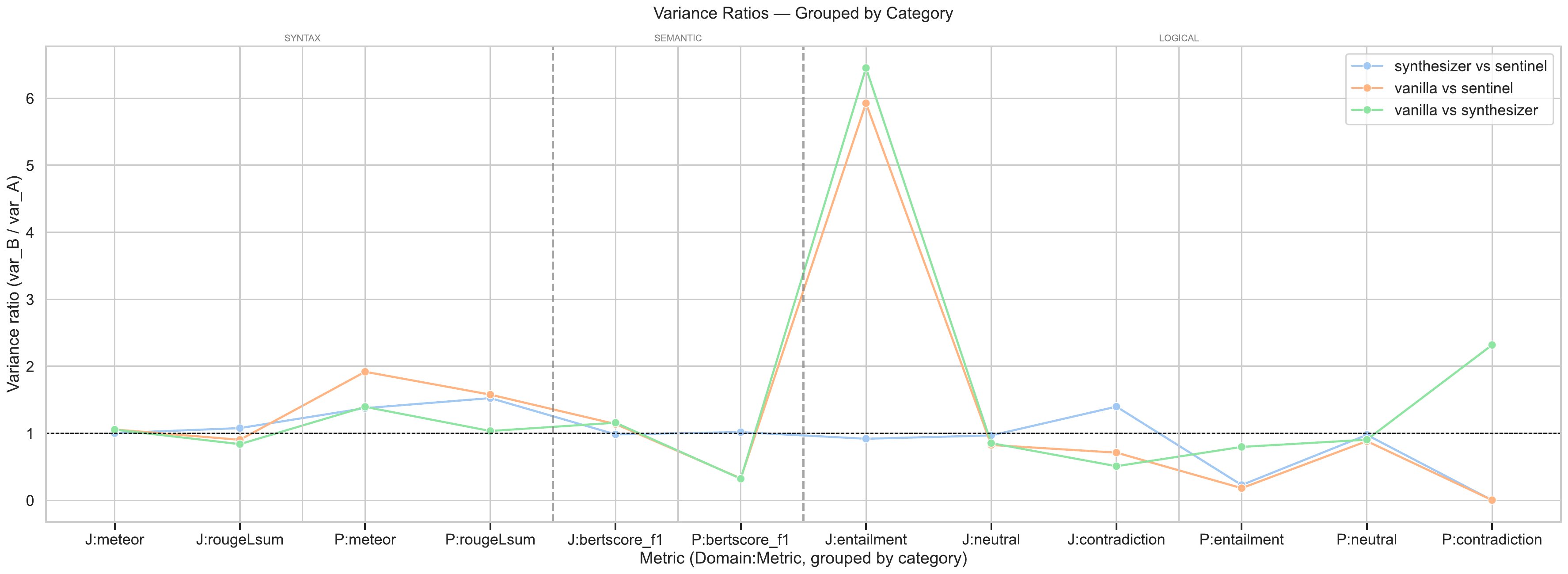}
		\caption{Variance ratio}
	\end{subfigure}
    \begin{subfigure}{0.8\linewidth}
		\includegraphics[width=\linewidth]{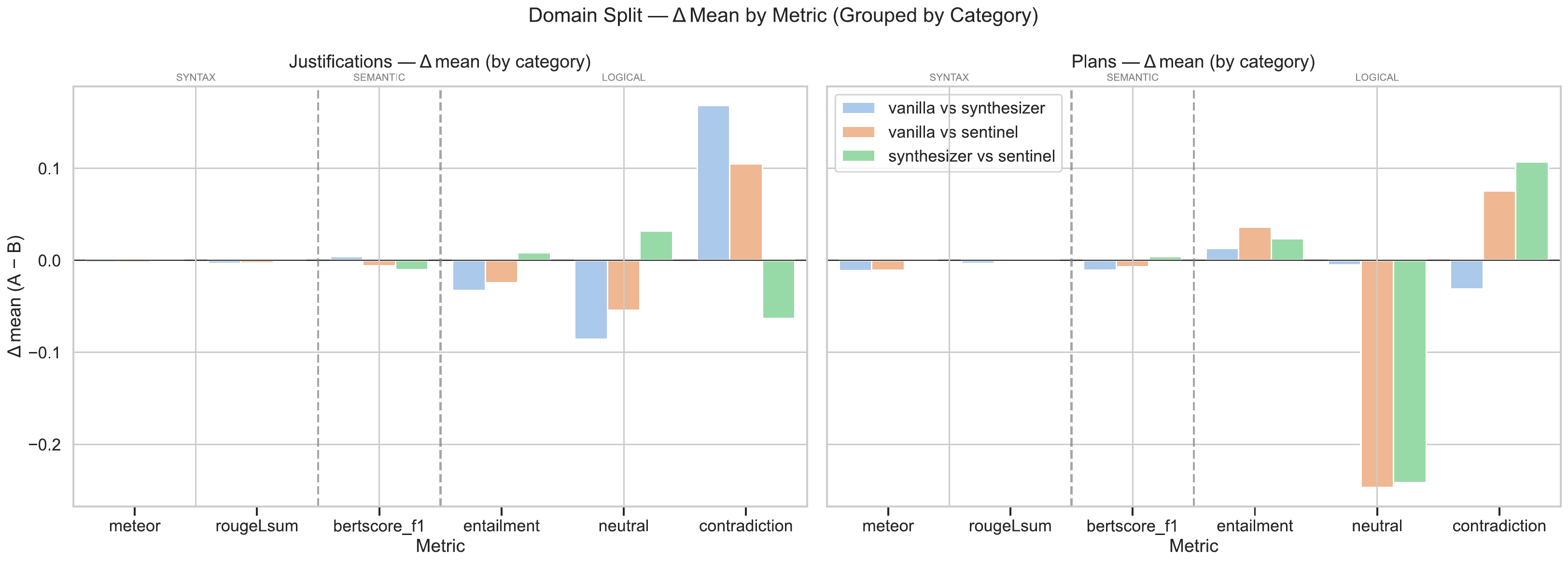}
		\caption{Delta mean between planners}
	\end{subfigure}
    \caption{Statistical analysis of planner performance}
	\label{fig:statistical_analysis}
\end{figure}
The BDD-X annotations, while useful, are often incomplete descriptions of a scene. For instance, an annotation might state the reason for an action is a "red light" while failing to mention a delivery van that is also very important for a safe plan. The model, by seeing the entire context, is forced to create a more complete plan that naturally is different from the less-detailed ground truth. This difference is therefore not a model failure but a result of it being more complete than the annotation itself, which makes the dataset an unreliable guide for evaluation.

Second, is a potential overuse of current evaluation measurements. Standard NLP benchmarks like BERTScore, ROUGE, and even NLI are not designed to measure the causal or time-based continuity of a plan. They are good at scoring static meaning overlap or logical connection, but are not sensitive to the coherence of a sequence. A planner could produce two individually good but contextually mismatched plans, and these measurements would fail to penalize the bad transition. They effectively measure the quality of the frames, not the film.

This finding strongly motivates the creation of new evaluation measurements that are able to capture coherence over time, a direction we explore in our conclusion.
\vspace{-10pt}
\vspace{-20pt}
\section{Conclusion and Future Recommendations}\label{sec6}
This work was motivated by the challenge of giving agentic planners a sense of continuous time, a key weakness in current systems that see the world in isolated frames. To investigate if language-based guidance could fix this, we designed a set of three planners (Static, Sentinel, and Synthesizer), where each one represents a deeper model of time integration. The later tests, conducted on curated subsets of the BDD-X dataset, showed that while time-based prompts changed how the planners reasoned, they did not yield statistically important gains in semantic or logical correctness. The study began with a question about continuity and ends with a clearer view of its limits and promise.

The principal finding that while temporal prompting reshapes reasoning style, it offers limited and inconsistent improvements in quantitative metrics. Simple time-awareness, as shown by the Sentinel planner, proved to be strong; it maintained semantic correctness that was comparable to the non-temporal baseline while using less computing power. In contrast, the deep time integration of the Synthesizer risked "compression drift," where over-summarization of historical context sometimes lowered the plan quality. Across all tests, the principle of not being worse than the baseline was followed, with performance differences for the time-aware planners remaining within a small margin ($\delta \approx 0.01–0.02$) of the baseline. Time-awareness thus appears as a measurable property of reasoning, not a guaranteed improvement.

Theoretically, this work argues for treating time as a first-class variable in agentic reasoning, bridging the gap between temporally-aware perception and high-level planning. The null statistical findings are not a rejection of the hypothesis but are instead a valuable description of its limits. By demonstrating the limitations of simply adding prompts together, this study highlights that understanding failure modes is a form of progress, which can guide future work toward better systems for memory and causality. Thinking about time thus becomes a special way to view how language models think in a sequence.

Based on these findings, we recommend three main directions for future
research. First, work should move from simple prompt joining to creating clear,
hidden memory modules. Architectures that learn to compress and retrieve
time-based information may overcome the context window and compression drift
limits seen here. Second, the trigger for time-based actions should change from
random sampling to a smart policy, where thinking is driven by changes in
meaning or prediction errors, which would use resources more efficiently.
Finally, these planners must be put back into an end-to-end system for
real-world testing. Validating the generated plans through a behavior-tree
controller in a simulated environment is critical to confirm that text-based
correctness leads to effective driving behavior, while also revealing the
practical constraints imposed by the currently high generation time. Reducing
this latency is essential for any meaningful deployment.


\bibliographystyle{style/IEEEtran}
\bibliography{refs/references}
\end{document}